\pdfoutput=1
\documentclass[11pt]{article}

\usepackage[final]{acl}

\usepackage{times}
\usepackage{latexsym}

\usepackage[T1]{fontenc}
\usepackage[utf8]{inputenc}

\usepackage{microtype}

\usepackage{inconsolata}

\usepackage{graphicx}

\title{Cool-Fusion: Fuse Large Language Models without Training}

\author{Cong Liu, Xiaojun Quan, Yan Pan, Weigang Wu, Xu Chen, Liang Lin \\
  Sun Yat-sen University \\
  \texttt{liucong3@mail.sysu.edu.cn, quanxj3@mail.sysu.edu.cn,  } \\ 
  \texttt{panyan5@mail.sysu.edu.cn, wuweig@mail.sysu.edu.cn} \\
  \texttt{chenxu35@mail.sysu.edu.cn, linliang@ieee.org}
  }

\begin{document}
\maketitle
\begin{abstract}
We focus on the problem of fusing two or more heterogeneous large language models (LLMs) to leverage their complementary strengths. One of the challenges of model fusion is high computational load, specifically in fine-tuning or aligning vocabularies. To address this, we propose Cool-Fusion, a simple yet effective approach that fuses the knowledge of source LLMs, which does not require training. Unlike ensemble methods, Cool-Fusion is applicable to any set of source LLMs that have different vocabularies. To overcome the vocabulary discrepancies among LLMs, we ensemble LLMs on text level, allowing them to rerank the generated texts by each other with different granularities. Extensive experiments have been conducted across a variety of benchmark datasets. On GSM8K, Cool-Fusion increases accuracy from three strong source LLMs by a significant margin of 17.4\%. 
\end{abstract}

\section{Introduction}

Different large language models (LLMs) exhibit diverse strengths and weaknesses due to various factors, such as datasets used for pre-training and fine-tuning, architectures, optimizers, hyperparameters, and training methodologies. Recent work \cite{Blender} has found that it is possible to develop fusion methods to harness the complementary potentials of the LLMs for improved general or task-specific performance, such as higher accuracy and better alignment with human preferences.

However, conventional ensemble approaches require the source LLMs to have the same token vocabulary, while weight merging \cite{soups, PAPA} is further limited to models with identical architectures. Although model fusion \cite{survey} has attracted increasing interest, it faces a series of challenges, including the formidable computational costs associated with training \cite{Composition, Bridging}, fine-tuning \cite{Blender}, distillation \cite{alpaca, wan2024knowledge, wan2024fusechat}, and the combinatorial optimization needed for vocabulary alignment \cite{wan2024knowledge, wan2024fusechat, Specializing, Bridging}. Therefore, existing fusion approaches are daunting for researchers and practitioners who cannot afford to train or fine-tuning LLMs, and are unsuitable for application scenarios that require rapid deployment.

Aiming for a general LLM fusion approach that is applicable to any set of source LLMs with diverse tokenizers, and is both cost-effective and fast to deploy, we propose Cool-Fusion to fuse the knowledge of heterogeneous LLMs without any training. The core of our algorithm combines the source LLMs to rerank text segments that they generate individually, rather than using the ensemble of LLMs as token generators with their own sets of disjoint vocabularies. In Cool-Fusion, we propose to fuse knowledge at text segments of different granularities, and discuss their pros and cons. An overview of Cool-Fusion is shown in Figure~\ref{fig:overview}. In summary, Cool-Fusion has the following properties:

\begin{figure*}[t]
 \centering
  \includegraphics[trim=63 328 238 162,clip,width=0.9\textwidth]{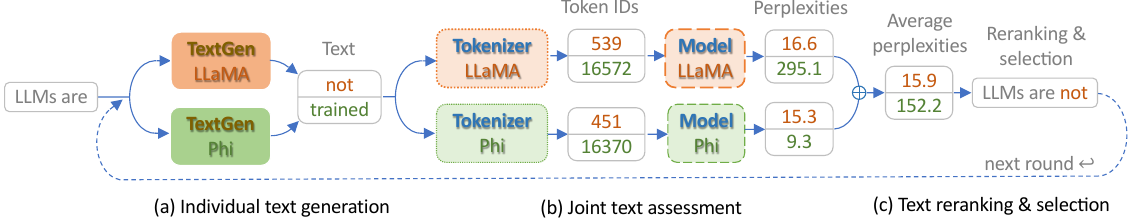}
  \caption{An illustration of Cool-Fusion. The TextGen component is illustrated in Figure~\ref{fig:textgen}.}
  \label{fig:overview}
\end{figure*}

\begin{table*}
  \centering
  \scriptsize
  \begin{tabular}{llllllllllll}
    \hline
    \textbf{\#iteration} & \textbf{0} & \textbf{1} & \textbf{2} & \textbf{3} & \textbf{4} & \textbf{5} & \textbf{6} & \textbf{7} & \textbf{8} & \textbf{9} & \textbf{10} \\
\hline
LLaMA-3 & \textbf{\_not} & \textbf{\_the} & \textbf{\_only} & \textbf{\_ones} & \_who & \textbf{\_can} & \textbf{\_be} & \textbf{\_used} & \textbf{\_for} & \textbf{\_this} & \textbf{\_purpose} \\
Phi-3 & \_trained & \_inherently & \_only & \_ones & \textbf{\_that} & \_can & \_be & \_used & \_for & \_this & \_purpose \\
\hline
LLaMA-3 & \_not & \_the & \_only & \_ones & \_who & \_have & \_been & \_affected & \_by & \_the & \_pandemic \\
Phi-3 & \_trained & \_on & \_vast & \_datasets & \_that & \_include & \_a & \_wide & \_variety & \_of & \_human \\
\hline
    \end{tabular}
  \caption{ A example of Cool-Fusion for 10 iterations following the example in Figure~\ref{fig:overview}. The first two rows show the text segments predicted by LLaMA-3 and Phi-3 jointly in Cool-Fusion, where the winning text segments are in bold. We use underscores to represent whitespaces. For comparison, the last two rows are text segments predicted by LLaMA-3 and Phi-3 individually. }
  \label{tb:run}
\end{table*}

\begin{itemize}

\item Simplicity: Cool-Fusion is simple both in concept and for implementation. Unlike prior approaches, Cool-Fusion starts to generate texts as soon as we have the source LLMs, since no training of any type is required. Consequently, we do not need to worry about the problems associated with fine-tuning and training, such as overfitting the training distribution, insufficient hyper-parameter tuning, or loss of generalization ability \cite{Specializing}.

\item Availability: Based on pure inference, Cool-Fusion can be accessed by a larger range of budget-limited researchers and practitioners. 

\item Scalability: Cool-Fusion alternates between a generation and an evaluation step. Each of the two steps invoke the source LLMs independently, and small amount of texts and scores are gathered and scattered between the steps. Given $k$ GPUs, Cool-Fusion is scalable to $k$ source LLMs with constant delay.

\item Superior performance: Albeit being simple, Cool-Fusion exhibits competitive performances over strong baselines, persistent across a wide range of challenging tasks.
\end{itemize}

We evaluated extensively on greedy completion benchmarks across various domains, including math (GSM8K, multilingual GSM, and MATH), and Q\&A (CoQA, DROP, TriviaQA). We experimented on an array of open-source LLMs, including the most recent state-of-the-art LLMs, namely LLaMA-3 8B \cite{llama}, Phi-3 mini \cite{phi3}, and GLM-4 9B \cite{glm}. Our results demonstrate that Cool-Fusion significantly outperforms the individual source LLMs as well as recent LLM fusion methods that require training. On the GSM8K dataset, Cool-Fusion increases the prediction accuracy from the best-performing source LLM LLaMA-3 8B by a significant margin of 17.4\%.

\section{Cool-Fusion: Fuse LLMs without Fine-tuning}

Since the token vocabularies are usually different across LLMs, a token predicted by one LLM may not find a deterministic counterpart in another LLM. For instance, the common tokens between LLaMA-3 and Phi-3 account for only 6.4\% of their total tokens, and those between Phi-3 and GLM-4 account for only 7.5\%. Prior approaches \cite{wan2024knowledge, wan2024fusechat, Specializing} resort to heuristics to find similar tokens across token vocabularies, which introduces errors and requires heavy training due to the combinatorial optimization complexity. In ensemble approaches \cite{Pack}, the predicted distributions on heterogeneous token vocabularies are first individually mapped into distributions on a shared tokens-vocabulary, and the next jointly predicted token from the shared vocabulary is the one that has the largest sum of logit values from these distributions. Inspired by this, we generalize the element of predicting from a single token to predicting a short text segment containing one or more tokens that can be commonly decoded by all heterogeneous tokenizers, and the criteria from the sum of logit values to the averaged perplexities of the text segment obtained from the LLMs. With this new approach, we can avoid the computation and inaccuracies associated with the mapping from individual token vocabularies into a single shared token vocabulary.

\subsection{Overview of Cool-Fusion} \label{sec:overview}

A text generation task involves generating a continuation of a given context. Our approach can be easily explained with a real running example, as illustrated in Figure~\ref{fig:overview}. Cool-Fusion features a text generation loop, where a text segment is generated at each iteration of the loop. In the example, we fuse two source LLMs, LLaMA-3 8B and Phi-3 mini, with the input context text being ``LLMs are''. Each iteration in the text generation loop consists of three steps: (1) each source LLM individually generates a text segment, (2) each source LLM computes a perplexity for every text segment generated in step 1, (3) the text segment with the smallest averaged perplexity is selected as the jointly predicted text segment, which is then broadcast to update all source LLMs. Next, we will discuss more details for each step, as illustrated in Figure~\ref{fig:overview}.

In step 1 of each iteration, a text generation component (TextGen) in each source LLM is responsible for generating text segments. Different implementations of TextGen may generate text segments of different lengths, ranging from minimal decodable text segments consisting of one or a few tokens to phrases containing several words. We will discuss two implementation options in Sections~\ref{sec:min} and \ref{sec:align}. In Figure~\ref{fig:overview}, the text segments generated by the two TextGen components are ``not'' and ``trained'', respectively. 

In step 2, each text segment is sent to all LLMs to obtain a perplexity using the key-value cache from the previous iteration, specifically the key-value cache before the generation of the text segment in the current iteration. Finally, the perplexities of each text segment are gathered from every LLM and are averaged to evaluate the text segment. In Figure~\ref{fig:overview}, the text segment ``not'' is first encoded by the tokenizers of LLaMA-3 8B and Phi-3 mini into token sequences [539,] and [451,]. These two token sequences are forwarded through their corresponding LLMs, resulting in two perplexities, 16.6 and 15.3, for text segment ``not'', which are finally averaged to 15.9. For better efficiency in this step, we forward all text segments, i.e. ``not'' and ``trained'', together in a batch through all LLMs.

In step 3, the winner among the text segments is selected based on their average perplexity computed in step 2. In Figure~\ref{fig:overview}, the winner is ``not'', whose average perplexity of 15.9 is better (smaller) than that of ``trained'', which is 152.2. We justify the adoption of average perplexity with two perspectives: the ensemble perspective and the critic perspective. From the ensemble perspective, the average perplexity is aligned with the cross-entropy objective of the ensemble of the LLMs. From the critic perspective, the LLMs leverage their complementary critical abilities to detect non-factual text segments by giving them high perplexities. Finally, the winning text segment is forwarded through all LLMs, except for the LLM that generated the winning text segment, to update their states before entering the next iteration. The winning text segment selected in our approach may not be optimal, and a natural improvement is to let each LLM generates its top-k text segments in step 1 using beam search.

\begin{figure}[]
 \centering
  \includegraphics[trim=73 153 553 291,clip,width=0.46\textwidth]{fig/overview}
  \caption{A contrived example illustrates our aligned text segments generation. In this example, the generated token sequence from LLaMA is first decoded into text, and then encoded and decoded by the tokenizers of two source LLMs into text segments: [``not'', ``the''] and [``no'', ``t'', ``the''], respectively. The aligned text segment ``not'' ends at the first common decodable boundary of all tokenizers, which helps to reduce biases in perplexity assessment due to the uneven text segment lengths across the tokenizers.}
  \label{fig:textgen}
\end{figure}

Table~\ref{tb:run} shows the running results of our Cool-Fusion following the example in Figure~\ref{fig:overview} with a side-by-side comparison of the generation from the two source LLMs. As we can see from this example, the text generated by Cool-Fusion is seldom identical to that of its source LLM, since the divergence accumulates from the different text segment in each iteration. It seems that shorter text segments can result in more flexibility and lower perplexity; however, this is not necessarily the case. We will present two options for the selection of text segment length in Sections~\ref{sec:min} and ~\ref{sec:align}. 

On the other extreme, the entire continuation can be used as a text segment, and sentence-level perplexity is employed to select (rerank) the the best continuation. In our Cool-Fusion approach, we can simultaneously employ an iterative fine-grained text segment selection and a coarse-grained sentence level reranking at the same time with almost no additional overhead. We let each source LLM independently predict a continuation segment in the same batch as each fine-grained text segment, with an additional overhead only on packing their key-value caches together. Then, we obtain $k$ individual continuations in addition to a jointly predicted continuation. Our experiment results show that reranking these $k+1$ continuations on their average perplexities can lead to substantial improvements over using the joint prediction alone.

\subsection{Shortest Text Segment} \label{sec:min}
We will discuss two implementations of the TextGen component, as shown in Figure~\ref{fig:textgen}. We prefer shorter text segments since they suggest finer-grained token selection and are therefore more likely to obtain a similar token sequence from an optimal token level ensemble approach. In this subsection, we will demonstrate how to generate the shortest possible text segments.

We define the shortest text segment as a text that can be decoded from the shortest token sequence generated by a greedy decoding process. Not all token sequences are decodable. For instance, LLaMA-2 uses three Unicode bytes as the tokens to encode a single Chinese character, so the first one or two of these tokens cannot be decoded. 

When decoding a token sequence into text, some tokenizers return additional information about the sequence of words in the text and the tokens that decode each of these words. In this case, we adopt the words as our shortest text segments since they are the minimal semantic units underlying a token sequence, although sometimes a word cannot be further divided into decodable subwords. 

Specifically, tokenizers from the LLaMA-3 tokenizer provide a \verb|word_ids| function that returns the IDs of the words in each decoded text, and a \verb|word_to_tokens| function that returns the indexes of the first and last tokens for each word id. Tokenizers derived from the LLaMA-2 tokenizer provide an \verb|offsets| property for each token, which contains the starting and ending character indexes in the text for the word decoded from the token.

For tokenizers that return decoded text without information about words, we derive shortest text segments as follows. Iteratively, we build a token sequence that initially contains only the next predicted tokens. Subsequently, a new next token is appended to the token sequence in each new iteration. In some iterations, if we can decode the current token sequence into a text that can be encoded back into the same token sequence, the decoded text is the shortest decodable text segment we need.

Table~\ref{tab:llms} lists the LLMs that we will use in our experiments and their categories according to the above discussion. 

\subsection{Aligned Text Segments} \label{sec:align}

% \begin{table}
%   \centering
%   \small
%   \begin{tabular}{lcccccccc}
%     \hline
%     \textbf{Metric} & \textbf{LLaMA-3} & \textbf{Phi-3} & \textbf{Cool$_{2}$} & \textbf{GLM-4} & \textbf{Rerank$_3$} & \textbf{Cool$_{-align}$} & \textbf{Cool} & \textbf{Cool+R}  \\
%     \hline
%     Avg. perplexity  & 1.48  & 1.35 & - & 1.46 & - & - &\textbf{1.29} & - \\
%     Accuracy  & 0.6914   & 0.6831 & 0.7233 & 0.6338 & 0.7779 & 0.7445 & 0.7468  & \textbf{0.812} \\
%     \hline
%     \end{tabular}
%   \caption{Averaged perplexities and accuracies in the GSM8K datasets (Section~\ref{sec:cmp}).}
%   \label{tab:cmp}
% \end{table}

\begin{table*}
  \centering
  \scriptsize
  \begin{tabular}{lllll}
    \hline
    \textbf{Name} & \textbf{Model ID} & \textbf{Parameters} & \textbf{Vocab size} & \textbf{Tokenizer category} \\
    \hline
   LLaMA-3 \cite{llama} & meta-llama/Meta-Llama-3-8B-Instruct & 8B & 128,256 & LLaMA-3 \\
    MPT \cite{mpt} & mosaicml/mpt-7b-instruct & 7B & 50,277 & LLaMA-3 \\
   LLaMA-2 \cite{llama} & meta-llama/Llama-2-7b-chat-hf & 7B & 32,000 & LLaMA-2 \\
    Phi-3  \cite{phi3}  & microsoft/Phi-3-mini-128k-instruct & 3.8B & 32,038 & LLaMA-2 \\
    OpenLLaMA \cite{openllama} & m-a-p/OpenLLaMA-Reproduce & 7B & 32,000 & LLaMA-2 \\
    GLM-4 \cite{glm2024chatglm} & THUDM/glm-4-9b-chat & 9B & 151,343 & OTHER \\
    ChatGLM-3 \cite{glm} & THUDM/chatglm3-6b & 6B & 64,796 & OTHER \\
    ChatGLM-2 \cite{glm} & THUDM/chatglm2-6b & 6B & 64,787 & OTHER \\
    Baichuan2 \cite{baichuan} & Baichuan2-7B-Chat & 7B & 125,696 & OTHER \\
    \hline
    \end{tabular}
  \caption{Source LLMs used in our experiments are divided into three categories in the last columns according to how to obtain their shortest text segments.}
  \label{tab:llms}
\end{table*}

\begin{table}
  \centering
  \scriptsize
  \begin{tabular}{lcccc}
    \hline
    \textbf{Metric} & \textbf{LLaMA-3} & \textbf{Phi-3} & \textbf{Cool$_{2}$} & \textbf{GLM-4}  \\
    \hline
    Avg. perplexity  & 1.48  & 1.35 & - & 1.46 \\
    Accuracy  & 0.6914   & 0.6831 & 0.7233 & 0.6338 \\
    \hline
     & \textbf{Rerank$_3$} & \textbf{Cool$_{-align}$} & \textbf{Cool} & \textbf{Cool+R}  \\
    \hline
    Avg. perplexity  & - & - &\textbf{1.29} & - \\
    Accuracy  & 0.7779 & 0.7445 & 0.7468  & \textbf{0.812} \\
    \hline
    \end{tabular}
  \caption{Averaged perplexities and accuracies in the GSM8K datasets (Section~\ref{sec:cmp}).}
  \label{tab:cmp}
\end{table}

In this subsection, we propose a better type of text segment to reduce the bias in the average perplexity used to select the best text segments generated by the source LLMs.

Different tokenizers may produce their shortest text segments of varying lengths. For instance, the text ``Multi-tasking'' is divided by LLaMA-3 and LLaMA-2 tokenizers into words [``Multi'', ``-tasking''] and [``Multi-tasking''], respectively.

On the other hand, perplexity, as a measure of how well a given model generates a continuation given a context, is the most widely used metric for evaluating language models due to simplicity and its alignment to the cross-entropy (CE) loss used for the next-token prediction objective. Since the latter confers multi-step reasoning ability to LLMs, we believe that perplexity is not only a measure of language fluency but also an indicator of inference correctness to some extent. The perplexity (PPL) of a token sequence $s$ is proportional to the average of the logits of the tokens:
\begin{equation}
PPL_u(s) = exp(\frac{1}{|s|}\sum_{s_i \in s} -\log p_u(s_i)),
\label{eq:ppl}
\end{equation}
where $\log p_u(s_i)$ is the logit output of the LM $u$ for each token $s_i$. 

However, as a measure of uncertainty, $-\log p_u(s_i)$ tends to be larger at the first token of each word. This is comparable to the observation on larger scales that the perplexity of the first word in a sentence is usually larger than those of the words following it, and that the perplexity of the first sentence in a paragraph is larger than those of the sentences following it. Here is a concrete example: in the LLaMA-3 8B model, the $-\log p_u(s_i)$'s values for the three tokens [``Multi'', ``-task'', ``ing''] are -2.66, -9.13, -14.69, respectively. Clearly, the model is more uncertain about the first token, and is more confident about the second token ``-task'' given the first token being ``Multi''.

Therefore, using perplexity as an assessment will bias towards longer text segments, and towards LLMs with tokenizers that generate text segments of larger average lengths. Suppose both LLaMA-3 and LLaMA-2 will predict the word ``Multi-tasking'', and their next text segments should be tied. Based on perplexity, however, the text segment ``Multi-'' from LLaMA-3 (-2.66) is regarded as worse than ``Multi-tasking'' from LLaMA-2 ($(-2.66-9.13-14.69)/3=-8.83$).

To mitigate this problem, we must reduce the discrepancies between the average lengths of the text segments generated by different source LLMs. To this end, we define a new aligned text segment for each source LLM as the shortest text segment that is generated by the LLM and is decodable by the tokenizers of all source LLMs.

\subsection{Incremental Encoding \& Decoding}

Both shortest text segments and aligned text segments require more frequent invocations of tokenizers than conventional decoding. In this subsection, we investigate an implementation issue about how to make decoding and encoding more efficient. First, let us examine the problem that not all tokenizers encode and decode incrementally, that is, the next text cannot be decoded solely from the next tokens, which results in significant delays as the encoding/decoding sequence increases.

It is expected that the text input and the tokenized sequence are reversibly convertible. For multilingual tokenizers, whitespace is treated as a normal symbol and preserved in the segmented tokens, allowing us to de-tokenize text without relying on language-specific rules such as: there is whitespace between two English words, but not between Chinese and Japanese words.

An instance of non-incremental encoding and decoding is the LLaMA-2 tokenizer, whose \verb|encode| and \verb|decode| functions are context-dependent and require complete token sequences or text to work correctly. For example, the \verb|decode| function in LLaMA-2 decodes the token [839] into ``If'' or ``~~If'' (with a preceeding space) depending on whether or not the token is the first token in the token sequence. Therefore, we cannot encode a new token in isolation, and the conventional method to decode a few new tokens is to encode the concatenation of all previous tokens and the new tokens, which makes it inefficient for long sequences.

To enable incremental decoding, we only prepend the tokens belonging to the previous $k$ decoded words to the new tokens, and we remove these $k$ words from the decoded text after decoding. We handle incremental encoding similarly by prepending $k$ decoded words to new text to be encode. Thus, we can encode and decode with constant computational complexity regardless of the context length. We empirically found that $k=4$ ensures correctness for both incremental encoding and decoding.

\section{Experiments}

Our experiments are conducted in a challenging scenario for LLM fusion, where the tokenizers of the source LLMs have very different token vocabularies and define text segments differently. A wide range of datasets is used to make our evaluations comprehensive. The questions that we want to answer from our experiments include: How does Cool-Fusion's performance compare with recent work? What are the contributions of its individual components, such as fine-grained perplexity-based text segment selection, shortest text segments, and aligned text segments? Is it a general method that performs well in various domains? Can it improve multilingual performance? Does its performance persist when fusing different LLMs?

% \begin{table}
%   \centering
%   \scriptsize
%   \begin{tabular}{ll}
%     \hline
%     \textbf{Method} & \textbf{GSM8K} \\
% \hline
% LLaMA2-7B-Chat & 24.64 \\
% ChatGLM2-6B & 30.78 \\
% Baichuan2-7B-Chat & 29.95 \\
% InternLM-7B-Chat & 32.30 \\
% TigerBot-7B-Chat-V3 & 27.29 \\
% Vicuna-7B-V1.5 & 18.88 \\
% ChineseAlpaca2-7B & 13.12 \\
% \hline
% MBR \cite{mbr} & 36.47(+4.17) \\
% PairRanker \cite{pairranker} & 39.58(+7.28) \\
% LLM-Blender \cite{Blender} & 34.80(+2.50) \\
% EVA \cite{Bridging} & 42.91(+10.61) \\
% \hline
% \hline
% LLaMA2-7B-Chat & 19.3 \\
% ChatGLM2-6B & 25.9 \\
% Baichuan2-7B-Chat & 26.9 \\
% \hline
% Cool (training-free, 3 source LLMs) & 33.5 (+6.6) \\
% \hline
%     \end{tabular}
%   \caption{We compare our results with recent model fusion algorithms that use training. Data in the first two blocks are from \cite{Bridging}, and those in the last two blocks are our results. Please note that this is not an apple-to-apple comparison: (1) we are comparing a training-free method to those that require different types of training, (2) the results of our Cool-Fusion are based on fusing three source LLMs due to our resource limitations, and (3) the scores of our source LLMs are on average more than 4 points lower than those reported in \cite{Bridging} due to differences in experimental settings.}
%   \label{tab:cmp1}
% \end{table}

\begin{table}
  \centering
  \scriptsize
  \begin{tabular}{llll}
    \hline
    \textbf{Method} & \textbf{src LLMs} & \textbf{Training} & \textbf{GSM8K} \\
\hline
LLaMA2-7B-Chat & - & - & 24.64 \\
ChatGLM2-6B & - & - & 30.78 \\
Baichuan2-7B-Chat & - & - & 29.95 \\
InternLM-7B-Chat & - & - & 32.30 \\
TigerBot-7B-Chat-V3 & - & - & 27.29 \\
Vicuna-7B-V1.5 & - & - & 18.88 \\
ChineseAlpaca2-7B & - & - & 13.12 \\
\hline
MBR & 7 above & - & 36.47 (+4.17) \\
PairRanker & 7 above & Ranker & 39.58 (+7.28) \\
LLM-Blender & 7 above & Merger & 34.80 (+2.50) \\
EVA & 7 above & Vocab Map & 42.91(+10.61) \\
\hline
\hline
LLaMA2-7B-Chat & - & - & 19.3 \\
ChatGLM2-6B & - & - & 25.9 \\
Baichuan2-7B-Chat & - & - & 26.9 \\
\hline
Cool & 3 above & Training-free & 33.5 (+6.6) \\
\hline
    \end{tabular}
  \caption{We compare our results with recent model fusion algorithms that use training. Data in the first two blocks are from \cite{Bridging}, and those in the last two blocks are our results. Please note that this is not an apple-to-apple comparison: (1) we are comparing a training-free method to those that require different types of training: PairRanker \cite{pairranker}, LLM-Blender \cite{Blender}, EVA \cite{Bridging}, (2) the results of our Cool-Fusion are based on fusing three source LLMs due to our resource limitations, and (3) the scores of our source LLMs are on average more than 4 points lower than those reported in \cite{Bridging} due to differences in experimental settings.}
  \label{tab:cmp1}
\end{table}

\subsection{Settings and Datasets}

We conduct experiments with several recent state-of-the-art open-source LLMs as our source LLMs, as listed in Table~\ref{tab:llms}.

To assess the performance of Cool-Fusion, we conduct experiments using the LM-Evaluation-Harness \cite{eval}, a benchmark framework designed to evaluate LLMs’ few-shot capabilities across various domains. We use its default settings, except for employing 3-shot prompting in all experiments. We conducted experiments on the following greedy text generation tasks.

\textbf{CoQA} \cite{coqa} requires understanding a text passage and answer a series of interconnected questions that appear in a conversation.

\textbf{DROP} \cite{drop} is a crowdsourced, adversarially-created, 96k-question benchmark, in which a system must resolve references in a question, perhaps to multiple input positions, and perform discrete operations over them (such as addition, counting, or sorting).

\textbf{TriviaQA} \cite{triviaqa} is challenging as the answers for a question may not be directly obtained by span prediction and the context is long.

\textbf{MATH} \cite{MATH} is a dataset of 12,500 challenging competition mathematics problems. Each problem in MATH has a full step-by-step answer derivations and explanations.

\textbf{GSM8K} \cite{gsm8k} is a dataset of high quality linguistically diverse grade school math word problems, that take between 2 and 8 steps of elementary calculations ($+ - \times \div$) to solve.

\textbf{MGSM} \cite{mgsm} stands for Multilingual Grade School Math Benchmark, where the same 250 problems from GSM8K are each translated in 10 languages other than English.

\textbf{Unscramble} \cite{unscramble} contains several tasks that are used for evaluating language models' ability to handle text manipulation tasks. The model must reconstruct the original word from scrambled letters. For example, in \emph{cyclle letters}, ``lyinevitab'' is reconstructed as ``inevitably'', in \emph{anagrams} where all but first and last or last 2 characters are scrambled, ``criroptuon'' is reconstructed as ``corruption'', and in \emph{random insertion} ``s.u!c/c!e.s s i/o/n'' s reconstructed as ``succession''.

\begin{table}[t]
  \centering
  \scriptsize
  \begin{tabular}{lcc}
    \hline
    \textbf{Method} & \textbf{Training} & \textbf{GSM8K} \\
    \hline    
    FuseLLM-7B [*] & Yes, via distillation & 13.8 \\
    Cool (ours) & No & 12.3 \\
    \hline
  \end{tabular}
  \caption{Comparison on the GSM8K dataset. Both methods use source LLMs: LLaMA2-7B-Chat \cite{llama}, MPT 7B \cite{mpt}, and OpenLLaMA-7B \cite{openllama}.}
  \label{tab:cmp2}
\end{table}

\subsection{Ablation study}

We compare Cool-Fusion with several source LLMs as baselines in Table~\ref{tab:cmp}. Cool$_2$, which fuses LLaMA-3 and Phi-3, immediately increases accuracy by 4.6\% and 5.9\%, respectively. Cool, which fuses LLaMA-3, Phi-3, and GLM-4, further increases the increments to 8.0\%, 9.3\%, and 17.8\%, respectively. This verifies the effectiveness of our fine-grained perplexity-based reranking.

$Cool_{-align}$ is an implementation using shortest text segment (Section~\ref{sec:min}), while $Cool$ implements aligned text segment (Section~\ref{sec:align}). $Cool_{-align}$ leads to a 0.3\% relative decrement, suffered from occasional bias in perplexity assessment.

Rerank is a simple reranking method, where each source LLM predicts a continuation individually, and these continuations are reranked using their average perplexities from all source LLMs. Rerank turns out to be very effective and it obtains a 12.5\% increment over LLaMA-3. Cool+R is a combination of Cool-Fusion and Rerank, which achieves a significant accuracy improvement of 17.4\% over LLaMA-3 and 4.4\% over Rerank.

\subsection{Compare with Other LLM Fusion Methods.}

The baseline results in Table 4 on GSM8K are directly sourced from Xu et al. (2024), as noted in the caption. Due to computational constraints, we limited our experiments to 3 LLMs and cannot reproduce those experiments that require various types of heavy computations. Here, MBR refers to minimum Bayes risk \cite{mbr}.

Due to differences in experimental setting, our scores for the source LLMs are, on average, 4 points lower than those reported in \cite{Bridging}. Table~\ref{tab:cmp1} shows that, although using only the three source LLMs and requiring no training, our Cool-Fusion reports a comparable score increment to existing methods that require different types of training to fuse all of the seven source LLMs. 

The results in Table~\ref{tab:cmp2} demonstrate that Cool achieves competitive performance (12.3) without requiring any training, while FuseLLM-7B, which relies on distillation, achieves only a marginally higher accuracy (13.8). 

These results underscores the efficiency and practicality of our training-free approach across different sets of models, making it a compelling alternative to resource-intensive methods.

% \begin{table}[t]
%   \centering
%   \small
%   \begin{tabular}{lcccc}
%     \hline
%     \textbf{Dataset} & \textbf{LLaMA-3} & \textbf{Phi-3} & \textbf{GLM-4} & \textbf{Cool} \\
%     \hline
%     Algebra & 0.2797 & 0.3783 & 0.1137 & \textbf{0.4330} \\
%     Count-prob & 0.1899 & 0.27 & 0.1983 & \textbf{0.2785} \\
%     Prealgebra  & 0.3846 & 0.4294 & 0.2939 & \textbf{0.5718} \\
%     % Geometry & 0.1357 & \textbf{0.2213} & 0.0752 & 0.1941 \\
%     % Inter algebra  & 0.0332 & 0.0487 & 0.0410 &  \\
%     % Num theory  & 0.1407 & 0.1519 & 0.0889 &  \\
%     % Precalc  & 0.033 & 0.0513 & 0.0403 &  \\
%     \emph{Average} & 0.2847 & 0.3592 & 0.2020 & \textbf{0.4278}\\
%     \hline
%     \end{tabular}
%   \caption{Accuracies in the Math dataset.}
%   \label{tab:math}
% \end{table}

\begin{figure}[t]
  \includegraphics[trim=60 595 280 100,clip,width=0.45\textwidth]{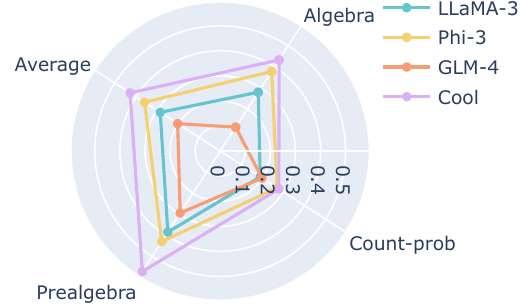}
  \caption{Accuracies in the Math dataset.}
  \label{fig:math}
\end{figure}

\subsection{Cross-domain Performances} \label{sec:Comparison} \label{sec:cmp}

\begin{figure}[t]
\flushright
  \includegraphics[trim=58 572 280 100,clip,width=0.42\textwidth]{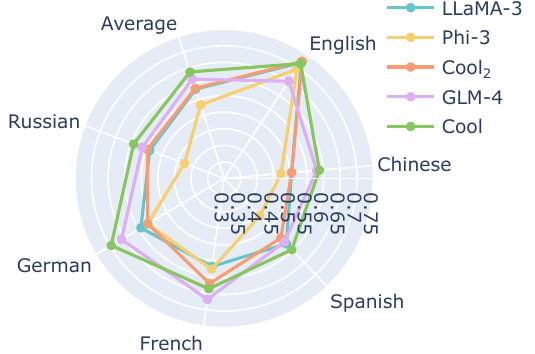}
  \caption{Accuracies in the multilingual GSM datasets.}
  \label{fig:lang}
\end{figure}

\begin{figure}[t]
  \includegraphics[trim=40 586 280 100,clip,width=0.45\textwidth]{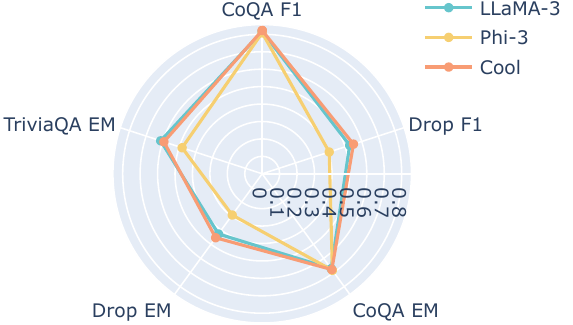}
  \caption{F1 \& EM in the four Q\&A datasets.}
  \label{fig:qa}
\end{figure}

\begin{figure}[t]
  \includegraphics[trim=40 606 280 100,clip,width=0.45\textwidth]{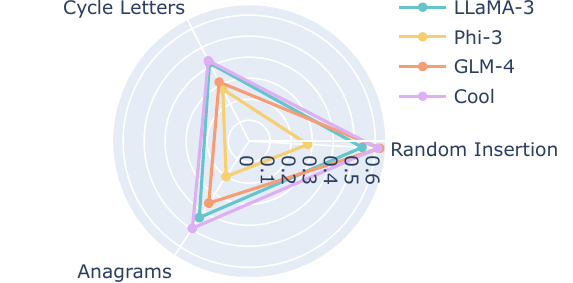}
  \caption{Accuracies in the Unscramble dataset.}
  \label{fig:unscramble}
\end{figure}

Next, we examine the general performance of Cool-Fusion in three different domains, where not all of source LLMs used have good performance. On the Q\&A datasets (Figure~\ref{fig:qa}), LLaMA-3 performs best, but GLM-4 fails to follow the output format in our 3-shot prompts. On the other hand, in the multilingual GSM datasets (Figure~\ref{fig:lang}), the overall performance of GLM-4 is the best, while Phi-3 does not perform well on multilingual data \cite{phi3}. Finally, on the Math dataset (Figure~\ref{fig:math}) and the Unscramble dataset (Figure~\ref{fig:unscramble}), Phi-3 is the best performer, and the other two LLMs lag behind with significant gaps. It is therefore challenging to fuse LLMs in these datasets where the performance of the source LLMs differs and fluctuates dramatically. Cool-Fusion either outperforms all source LLMs or is comparable to the best performer and not being affected by the poorer ones, which shows that Cool-Fusion is a stable method for fusing source LLMs across different domains.

Comparing the performance of Cool-Fusion in Table~\ref{tab:cmp} with that in Figure~\ref{fig:math}, we can see that Cool-Fusion performs much better on GSM8K than on multilingual GSM, although the latter is a translated subset of the former. This is probably because multilingual GSM contains a larger proportion of hard problems than in GSM8K. 

\subsection{Summary of Experiments}

In this section, we verify the effectiveness of the components in Cool-Fusion through ablation studies, which shows that our Cool-Fusion achieves significant improvements over the source LLMs on challenging tasks. It is able to achieve further advances when combined with other approaches, and persistently being better or comparable to the best-performing source LLMs even when some of them exhibit deteriorated performance. Our Cool-Fusion shows comparable performance with recent state-of-the-art LLM fusion methods that require training, and its performance persist when fusing different LLMs.

\section{Related Work}

Cool-Fusion aligns with established ensemble principles: (1) the Condorcet Jury Theorem \cite{Jury}, which justifies more independent models and (2) the bias-variance tradeoff, which suggests reduced variance with more models. In this section, we summarize prior work on model and LLM fusion. To our knowledge, prior work on fusion of heterogeneous LLMs involve different types of training.

% \begin{table}[t]
%   \centering
%   \scriptsize
%   \begin{tabular}{lccccc}
%     \hline
%     \textbf{Dataset} & \textbf{LLaMA-3} & \textbf{Phi-3} & \textbf{Cool$_{2}$} & \textbf{GLM-4} & \textbf{Cool} \\
%     \hline
%     English  &  0.72  & 0.696   & \textbf{0.724} &  0.652 & 0.716  \\
%     Chinese  &  0.504  &  0.472  & 0.504  &  0.58 & \textbf{0.588} \\
%     Spanish  &  0.572  &  0.456  & 0.548 &  0.564 & \textbf{0.596} \\
%     French  &  0.568  &  0.576  & 0.62 &  \textbf{0.668} &0.636 \\
%     German  &  0.592  &  0.568  & 0.568 & 0.66 & \textbf{0.696} \\
%     Russian  &  0.54 & 0.428 & 0.548 & 0.564 & \textbf{0.592} \\
%     % Japanese  &  0.4  & 0.316  & 0.376 & 0.584 &  \\
%     % Thai  & 0.4  & 0.056  & 0.352 & 0.432 & \\
%     % Swahili  & 0.296 & 0.06 & 0.196 & 0.372  & \\
%     % Bengali  & 0.228 & 0.012  & 0.112 & 0.384  & \\
%     % Telugu  & 0.06  & 0.016  & 0.012 & 0.112 & \\
%     \emph{Average} & 0.5827 & 0.5327 & 0.5853 & 0.6147 & \textbf{0.6373} \\
%     \hline
%     \end{tabular}
%   \caption{Accuracies in the multilingual GSM datasets.}
%   \label{tab:lang}
% \end{table}

\textbf{Reranking} methods first generate multiple candidates via probabilistic sampling, or by prompting LLMs. The quality of the candidates are then assessed using different scoring methods \cite{summareranker, Blender}.

\textbf{Alignment} matches the units of prediction from multiple models, i.e. the vocabularies of different LLMs. Since finding the optimal alignment is a combinatorial optimization problem, alignment between vocabulary is still an open problem. FuseLLM \cite{wan2024knowledge}, FuseChat \cite{wan2024fusechat}, and Specialized \cite{Specializing} use the edit-distance between tokens to map token distributions between LLMs, while EVA \cite{Bridging} trains a vocabulary projection matrix. 

However, it is unclear if the alignment approaches \cite{Pack, Bridging}, which assume substantial amount of common tokens across vocabularies, will work for Unicode vocabularies, where different tokenizers may share little portion of their symbols: Unicode bytes are the basic symbols in Phi-3 \cite{phi3}, while sub-word tokenization for Chinese \cite{subchar} uses glyph or pronunciation encoding.

% \begin{table}[t]
%   \centering
%   \small
%   \begin{tabular}{lcccc}
%     \hline
%     \textbf{Dataset} & \textbf{LLaMA-3} & \textbf{Phi-3} & \textbf{Cool} \\
%     \hline    
%     \textbf{F1}  &  &  &  \\
%     CoQA  & 0.8172  & 0.8091  & \textbf{0.8225}  \\
%     Drop & 0.5277 & 0.4041 & \textbf{0.5504} \\
%     \textit{Average} & 0.6724 & 0.6066 & \textbf{0.6865} \\
%     \hline
%     \textbf{EM}  &  &  &  \\
%     CoQA  & 0.6727 & \textbf{0.6848} & 0.6795 \\
%     Drop & 0.4268 & 0.2925 & \textbf{0.4539} \\
%     TriviaQA  & \textbf{0.6121} & 0.4829 & 0.5927 \\
%     % TruthfulQA-gen  &  &  &  \\
%     \textit{Average} & 0.5705 & 0.4867 & \textbf{0.5754} \\
%     \hline
%     \end{tabular}
%   \caption{F1 \& EM in the four Q\&A datasets.}
%   \label{tab:qa}
% \end{table}

\textbf{Ensembling} approaches conventionally require the source models to have the same token vocabulary, which can be partially relaxed by vocabulary alignment \cite{Pack}. LLM-Blender \cite{Blender} ensembles the outputs from several source LLMs by firstly using a fine-tuned ranking model to predict the top-ranked outputs, then it uses another fine-tuned LLM to generates a fused output. EVA \cite{Bridging} proposes to ensemble LLMs via a pre-trained vocabulary alignment matrix to enable a fine-grained token-level ensemble at each generation step.

\textbf{Weight average}. Researchers do not limit themselves to predictions, e.g. logics. Model soups \cite{soups}, which average the weights of multiple models fine-tuned with different hyper-parameter configurations, often improves accuracy and robustness. PAPA \cite{PAPA} obtains a strong single model by training a population of models and averaging them once-in-a-while or slowly pushing them toward the average. These methods require no training data, but the models to fuse must be of the same architecture.

% \begin{table}[t]
%   \centering
%   \scriptsize
%   \begin{tabular}{lcccc}
%     \hline
%     \textbf{Task} & \textbf{LLaMA-3 8B} & \textbf{Phi-3 Mini} & \textbf{GLM-4 9B} & \textbf{Cool} \\
%     \hline    
%     Anagrams  & 0.4341  & 0.2007  & 0.3517  & \textbf{0.4948}  \\
%     Cycle Letters & 0.4175 & 0.2777 & 0.3163 & \textbf{0.4297} \\
%     Random Insertion & 0.5382 & 0.2795 & 0.6228 & \textbf{0.6125} \\
%     \hline
%     \end{tabular}
%   \caption{Accuracies in the Unscramble dataset.}
%   \label{tab:unscramble}
% \end{table}

\textbf{Knowledge distillation}. Alpaca \cite{alpaca} used text-davinci-003 to generate the instruction data to distill a 7B LLaMA \cite{llama} model to reduce the cost of training LLMs from scratch. FuseLLM \cite{wan2024knowledge, wan2024fusechat} applies cost-effective distillation to merge pre-trained LLMs into a more potent model.

\textbf{Multi-agent} approaches enable an orchestration of a collection of LLM modules working together, each with different potentials. MetaGPT \cite{MetaGPT} encodes a standardized operating procedure (SOP) for software development into a prompt sequence. It breaks down complex tasks into subtasks, allowing agents with different domain expertise--such as architecture design and code debugging--to work harmoniously.

\textbf{Beam search} is importance for generation tasks like summarization and machine translation, and beam search with a single LLM often outperforms multi-LLM fusion on tasks like machine translation \cite{mbr} (e.g., MBR decoding achieves state-of-the-art results). This paper focuses on challenging generation tasks that require deep understanding and reasoning. Greedy generation was chosen for its simplicity and effectiveness, as it is widely used in practice and has been shown to perform well for large language models (LLMs).

\textbf{Others} Contrastive decoding \cite{Contrastive} exploits the contrasts between expert and amateur LLMs by choosing tokens that maximize their log-likelihood difference to amplify the good expert behavior and diminish the undesired amateur behavior. CALM \cite{Composition} introduces cross-attention between models to compose their representations and enable new capabilities.

\section{Conclusion and Future Directions}

In this work, we propose Cool-Fusion, a simple yet effective approach that fuses the knowledge of heterogeneous source LLMs. Extensive experiments with challenging datasets and strong source LLMs verify the persistent improvements and robustness of our proposal. 

Future work can focus on better measurements for text segments \cite{kang2025scalable}, improving inference speed, including streamlining different inference processes to fill GPU vacancies waiting for communication, parallelizing tokenizers to find out whether a text segment is decodable by all tokenizers, using longer text segments to reduce communication overhead between LLMs.

\section*{Limitations}

The inference speed of our implementation of Cool-Fusion is about six times slower than that of a standard LLM, mainly due to the additional communication among LLMs and the frequent invocation of tokenizers. Further optimizations, such as streamlining different inference processes or implementing parallel tokenizers, might increase the speed of Cool-Fusion.

Due to resource limitations, we only conduct experiments with two and three source LLMs. We used the automatic metrics that come with the Evaluation Harness \cite{eval}. Human or GPT-4 evaluations could provide us with more reliable and comprehensive results.

\section*{Ethical Statement}

This work fully complies with the ACL Ethics Policy. We declare that there are no ethical issues in this paper, to the best of our knowledge.

\bibliography{custom}

\begin{thebibliography}{34}
\providecommand{\natexlab}[1]{#1}

\bibitem[{Abdin et~al.(2024)Abdin, Ade~Jacobs, Awan, Aneja, Awadallah,
  Hassan~Awadalla, Bach, Bahree, Bakhtiari, Behl, Benhaim, Bilenko, Bjorck,
  Bubeck, Cai, Mendes, Chen, Chaudhary, Chopra, Giorno, de~Rosa, Dixon, Eldan,
  Iter, Goswami, Gunasekar, Haider, Hao, Hewett, Huynh, Javaheripi, Jin,
  Kauffmann, Karampatziakis, Kim, Khademi, Kurilenko, Lee, Lee, Li, Liang, Liu,
  Lin, Lin, Madan, Mitra, Modi, Nguyen, Norick, Patra, Perez-Becker, Portet,
  Pryzant, Qin, Radmilac, Rosset, Roy, Saarikivi, Saied, Salim, Santacroce,
  Shah, Shang, Sharma, Song, Ruwase, Wang, Ward, Wang, Witte, Wyatt, Xu, Xu,
  Xu, Yadav, Yang, Yang, Yu, Zhang, Zhang, Zhang, Zhang, Zhang, Zhang, and
  Zhou}]{phi3}
Marah~I Abdin, Sam Ade~Jacobs, Ammar~Ahmad Awan, Jyoti Aneja, Ahmed Awadallah,
  Hany Hassan~Awadalla, Nguyen Bach, Amit Bahree, Arash Bakhtiari, Harkirat
  Behl, Alon Benhaim, Misha Bilenko, Johan Bjorck, Sébastien Bubeck, Martin
  Cai, Caio César~Teodoro Mendes, Weizhu Chen, Vishrav Chaudhary, Parul
  Chopra, Allie~Del Giorno, Gustavo de~Rosa, Matthew Dixon, Ronen Eldan, Dan
  Iter, Abhishek Goswami, Suriya Gunasekar, Emman Haider, Junheng Hao,
  Russell~J. Hewett, Jamie Huynh, Mojan Javaheripi, Xin Jin, Piero Kauffmann,
  Nikos Karampatziakis, Dongwoo Kim, Mahmoud Khademi, Lev Kurilenko, James~R.
  Lee, Yin~Tat Lee, Yuanzhi Li, Chen Liang, Weishung Liu, Xihui~(Eric) Lin,
  Zeqi Lin, Piyush Madan, Arindam Mitra, Hardik Modi, Anh Nguyen, Brandon
  Norick, Barun Patra, Daniel Perez-Becker, Thomas Portet, Reid Pryzant, Heyang
  Qin, Marko Radmilac, Corby Rosset, Sambudha Roy, Olli Saarikivi, Amin Saied,
  Adil Salim, Michael Santacroce, Shital Shah, Ning Shang, Hiteshi Sharma, Xia
  Song, Olatunji Ruwase, Xin Wang, Rachel Ward, Guanhua Wang, Philipp Witte,
  Michael Wyatt, Can Xu, Jiahang Xu, Weijian Xu, Sonali Yadav, Fan Yang, Ziyi
  Yang, Donghan Yu, Chengruidong Zhang, Cyril Zhang, Jianwen Zhang, Li~Lyna
  Zhang, Yi~Zhang, Yunan Zhang, and Xiren Zhou. 2024.
\newblock \href {https://arxiv.org/abs/2404.14219} {Phi-3 technical report: A
  highly capable language model locally on your phone}.
\newblock \emph{Preprint}, arXiv:2404.14219.

\bibitem[{Baichuan.(2023)}]{baichuan}
Baichuan. 2023.
\newblock \href {https://arxiv.org/abs/2309.10305} {{Baichuan 2: Open
  large-scale lan- guage models.}}
\newblock \emph{Preprint}, arXiv:2309.10305.

\bibitem[{Bansal et~al.(2024)Bansal, Samanta, Dalmia, Gupta, Ganapathy, Bapna,
  Jain, and Talukdar}]{Composition}
Rachit Bansal, Bidisha Samanta, Siddharth Dalmia, Nitish Gupta, Sriram
  Ganapathy, Abhishek Bapna, Prateek Jain, and Partha Talukdar. 2024.
\newblock \href {https://openreview.net/forum?id=jjA4O1vJRz} {{LLM} augmented
  {LLM}s: Expanding capabilities through composition}.
\newblock In \emph{The Twelfth International Conference on Learning
  Representations}.

\bibitem[{Brown et~al.(2020)Brown, Mann, Ryder, Subbiah, Kaplan, Dhariwal,
  Neelakantan, Shyam, Sastry, Askell, Agarwal, Herbert-Voss, Krueger, Henighan,
  Child, Ramesh, Ziegler, Wu, Winter, Hesse, Chen, Sigler, Litwin, Gray, Chess,
  Clark, Berner, McCandlish, Radford, Sutskever, and Amodei}]{unscramble}
Tom Brown, Benjamin Mann, Nick Ryder, Melanie Subbiah, Jared~D Kaplan, Prafulla
  Dhariwal, Arvind Neelakantan, Pranav Shyam, Girish Sastry, Amanda Askell,
  Sandhini Agarwal, Ariel Herbert-Voss, Gretchen Krueger, Tom Henighan, Rewon
  Child, Aditya Ramesh, Daniel Ziegler, Jeffrey Wu, Clemens Winter, Chris
  Hesse, Mark Chen, Eric Sigler, Mateusz Litwin, Scott Gray, Benjamin Chess,
  Jack Clark, Christopher Berner, Sam McCandlish, Alec Radford, Ilya Sutskever,
  and Dario Amodei. 2020.
\newblock \href
  {https://proceedings.neurips.cc/paper/2020/file/1457c0d6bfcb4967418bfb8ac142f64a-Paper.pdf}
  {Language models are few-shot learners}.
\newblock In \emph{Advances in Neural Information Processing Systems},
  volume~33, pages 1877--1901. Curran Associates, Inc.

\bibitem[{Chen et~al.(2023)Chen, Zaharia, and Zou}]{pairranker}
Lingjiao Chen, Matei Zaharia, and James Zou. 2023.
\newblock \href {https://arxiv.org/abs/2305.05176} {{Frugalgpt: How to use
  large language models while reducing cost and improving performance}}.
\newblock \emph{Preprint}, arXiv:2305.05176.

\bibitem[{Cobbe et~al.(2021)Cobbe, Kosaraju, Bavarian, Chen, Jun, Kaiser,
  Plappert, Tworek, Hilton, Nakano, Hesse, and Schulman}]{gsm8k}
Karl Cobbe, Vineet Kosaraju, Mohammad Bavarian, Mark Chen, Heewoo Jun, Lukasz
  Kaiser, Matthias Plappert, Jerry Tworek, Jacob Hilton, Reiichiro Nakano,
  Christopher Hesse, and John Schulman. 2021.
\newblock Training verifiers to solve math word problems.
\newblock \emph{arXiv preprint arXiv:2110.14168}.

\bibitem[{de~Condorcet(1785)}]{Jury}
Marquis de~Condorcet. 1785.
\newblock Essai sur l'application de l'analyse à la probabilité des
  décisions rendues à la pluralité des voix (png).

\bibitem[{Dua et~al.(2019)Dua, Wang, Dasigi, Stanovsky, Singh, and
  Gardner}]{drop}
Dheeru Dua, Yizhong Wang, Pradeep Dasigi, Gabriel Stanovsky, Sameer Singh, and
  Matt Gardner. 2019.
\newblock \href {https://doi.org/10.18653/v1/N19-1246} {{DROP}: A reading
  comprehension benchmark requiring discrete reasoning over paragraphs}.
\newblock In \emph{Proceedings of the 2019 Conference of the North {A}merican
  Chapter of the Association for Computational Linguistics: Human Language
  Technologies, Volume 1 (Long and Short Papers)}, pages 2368--2378,
  Minneapolis, Minnesota. Association for Computational Linguistics.

\bibitem[{Farinhas and et~al.(2023)}]{mbr}
Antonio Farinhas and et~al. 2023.
\newblock \href {https://arxiv.org/abs/2310.11430} {{An empirical study of
  translation hypothesis ensembling with large language models}}.
\newblock \emph{Preprint}, arXiv:2310.11430.

\bibitem[{Fu et~al.(2023)Fu, Peng, Ou, Sabharwal, and Khot}]{Specializing}
Yao Fu, Hao Peng, Litu Ou, Ashish Sabharwal, and Tushar Khot. 2023.
\newblock Specializing smaller language models towards multi-step reasoning.
\newblock In \emph{Proceedings of the 40th International Conference on Machine
  Learning}, ICML'23. JMLR.org.

\bibitem[{Gao et~al.(2023)Gao, Tow, Abbasi, Biderman, Black, DiPofi, Foster,
  Golding, Hsu, Le~Noac'h, Li, McDonell, Muennighoff, Ociepa, Phang, Reynolds,
  Schoelkopf, Skowron, Sutawika, Tang, Thite, Wang, Wang, and Zou}]{eval}
Leo Gao, Jonathan Tow, Baber Abbasi, Stella Biderman, Sid Black, Anthony
  DiPofi, Charles Foster, Laurence Golding, Jeffrey Hsu, Alain Le~Noac'h,
  Haonan Li, Kyle McDonell, Niklas Muennighoff, Chris Ociepa, Jason Phang,
  Laria Reynolds, Hailey Schoelkopf, Aviya Skowron, Lintang Sutawika, Eric
  Tang, Anish Thite, Ben Wang, Kevin Wang, and Andy Zou. 2023.
\newblock \href {https://doi.org/10.5281/zenodo.10256836} {A framework for
  few-shot language model evaluation}.

\bibitem[{Geng and Liu(May 2023)}]{openllama}
Xinyang Geng and Hao Liu. May 2023.
\newblock Openllama: An open reproduction of llama.

\bibitem[{GLM(2024)}]{glm2024chatglm}
Team GLM. 2024.
\newblock \href {https://arxiv.org/abs/2406.12793} {Chatglm: A family of large
  language models from glm-130b to glm-4 all tools}.
\newblock \emph{Preprint}, arXiv:2406.12793.

\bibitem[{Hendrycks et~al.(2021)Hendrycks, Burns, Kadavath, Arora, Basart,
  Tang, Song, and Steinhardt}]{MATH}
Dan Hendrycks, Collin Burns, Saurav Kadavath, Akul Arora, Steven Basart, Eric
  Tang, Dawn Song, and Jacob Steinhardt. 2021.
\newblock \href
  {https://datasets-benchmarks-proceedings.neurips.cc/paper_files/paper/2021/file/be83ab3ecd0db773eb2dc1b0a17836a1-Paper-round2.pdf}
  {Measuring mathematical problem solving with the math dataset}.
\newblock In \emph{Proceedings of the Neural Information Processing Systems
  Track on Datasets and Benchmarks}, volume~1.

\bibitem[{Hong et~al.(2024)Hong, Zhuge, Chen, Zheng, Cheng, Wang, Zhang, Wang,
  Yau, Lin, Zhou, Ran, Xiao, Wu, and Schmidhuber}]{MetaGPT}
Sirui Hong, Mingchen Zhuge, Jonathan Chen, Xiawu Zheng, Yuheng Cheng, Jinlin
  Wang, Ceyao Zhang, Zili Wang, Steven Ka~Shing Yau, Zijuan Lin, Liyang Zhou,
  Chenyu Ran, Lingfeng Xiao, Chenglin Wu, and J{\"u}rgen Schmidhuber. 2024.
\newblock \href {https://openreview.net/forum?id=VtmBAGCN7o} {Meta{GPT}: Meta
  programming for a multi-agent collaborative framework}.
\newblock In \emph{The Twelfth International Conference on Learning
  Representations}.

\bibitem[{Jiang et~al.(2023)Jiang, Ren, and Lin}]{Blender}
Dongfu Jiang, Xiang Ren, and Bill~Yuchen Lin. 2023.
\newblock \href {https://doi.org/10.18653/v1/2023.acl-long.792} {{LLM}-blender:
  Ensembling large language models with pairwise ranking and generative
  fusion}.
\newblock In \emph{Proceedings of the 61st Annual Meeting of the Association
  for Computational Linguistics (Volume 1: Long Papers)}, pages 14165--14178,
  Toronto, Canada. Association for Computational Linguistics.

\bibitem[{Jolicoeur-Martineau et~al.(2024)Jolicoeur-Martineau, Gervais, FATRAS,
  Zhang, and Lacoste-Julien}]{PAPA}
Alexia Jolicoeur-Martineau, Emy Gervais, Kilian FATRAS, Yan Zhang, and Simon
  Lacoste-Julien. 2024.
\newblock \href {https://openreview.net/forum?id=cPDVjsOytS} {Population
  parameter averaging ({PAPA})}.
\newblock \emph{Transactions on Machine Learning Research}.

\bibitem[{Joshi et~al.(2017)Joshi, Choi, Weld, and Zettlemoyer}]{triviaqa}
Mandar Joshi, Eunsol Choi, Daniel Weld, and Luke Zettlemoyer. 2017.
\newblock \href {https://doi.org/10.18653/v1/P17-1147} {{T}rivia{QA}: A large
  scale distantly supervised challenge dataset for reading comprehension}.
\newblock In \emph{Proceedings of the 55th Annual Meeting of the Association
  for Computational Linguistics (Volume 1: Long Papers)}, pages 1601--1611,
  Vancouver, Canada. Association for Computational Linguistics.

\bibitem[{Kang et~al.(2025)Kang, Zhao, and Song}]{kang2025scalable}
Zhewei Kang, Xuandong Zhao, and Dawn Song. 2025.
\newblock Scalable best-of-n selection for large language models via
  self-certainty.
\newblock \emph{arXiv preprint arXiv:2502.18581}.

\bibitem[{Li et~al.(2023{\natexlab{a}})Li, Peng, Zhang, Ding, Hu, and
  Shen}]{survey}
Weishi Li, Yong Peng, Miao Zhang, Liang Ding, Han Hu, and Li~Shen.
  2023{\natexlab{a}}.
\newblock Deep model fusion: A survey.
\newblock \emph{arXiv preprint arXiv:2309.15698}.

\bibitem[{Li et~al.(2023{\natexlab{b}})Li, Holtzman, Fried, Liang, Eisner,
  Hashimoto, Zettlemoyer, and Lewis}]{Contrastive}
Xiang~Lisa Li, Ari Holtzman, Daniel Fried, Percy Liang, Jason Eisner, Tatsunori
  Hashimoto, Luke Zettlemoyer, and Mike Lewis. 2023{\natexlab{b}}.
\newblock \href {https://doi.org/10.18653/v1/2023.acl-long.687} {Contrastive
  decoding: Open-ended text generation as optimization}.
\newblock In \emph{Proceedings of the 61st Annual Meeting of the Association
  for Computational Linguistics (Volume 1: Long Papers)}, pages 12286--12312,
  Toronto, Canada. Association for Computational Linguistics.

\bibitem[{Mavromatis et~al.(2024)Mavromatis, Karypis, and Karypis}]{Pack}
Costas Mavromatis, Petros Karypis, and George Karypis. 2024.
\newblock \href {https://arxiv.org/abs/2404.11531} {Pack of llms: Model fusion
  at test-time via perplexity optimization}.
\newblock \emph{Preprint}, arXiv:2404.11531.

\bibitem[{Ravaut et~al.(2022)Ravaut, Joty, and Chen}]{summareranker}
Mathieu Ravaut, Shafiq Joty, and Nancy Chen. 2022.
\newblock \href {https://doi.org/10.18653/v1/2022.acl-long.309}
  {{S}umma{R}eranker: A multi-task mixture-of-experts re-ranking framework for
  abstractive summarization}.
\newblock In \emph{Proceedings of the 60th Annual Meeting of the Association
  for Computational Linguistics (Volume 1: Long Papers)}, pages 4504--4524,
  Dublin, Ireland. Association for Computational Linguistics.

\bibitem[{Reddy et~al.(2019)Reddy, Chen, and Manning}]{coqa}
Siva Reddy, Danqi Chen, and Christopher~D. Manning. 2019.
\newblock \href {https://doi.org/10.1162/tacl_a_00266} {{C}o{QA}: A
  conversational question answering challenge}.
\newblock \emph{Transactions of the Association for Computational Linguistics},
  7:249--266.

\bibitem[{Saparov and He(2023)}]{mgsm}
Abulhair Saparov and He~He. 2023.
\newblock \href {https://openreview.net/forum?id=qFVVBzXxR2V} {Language models
  are greedy reasoners: A systematic formal analysis of chain-of-thought}.
\newblock In \emph{The Eleventh International Conference on Learning
  Representations}.

\bibitem[{Si et~al.(2023)Si, Zhang, Chen, Qi, Wang, Liu, Wang, Liu, and
  Sun}]{subchar}
Chenglei Si, Zhengyan Zhang, Yingfa Chen, Fanchao Qi, Xiaozhi Wang, Zhiyuan
  Liu, Yasheng Wang, Qun Liu, and Maosong Sun. 2023.
\newblock \href {https://doi.org/10.1162/tacl_a_00560} {Sub-character
  tokenization for {C}hinese pretrained language models}.
\newblock \emph{Transactions of the Association for Computational Linguistics},
  11:469--487.

\bibitem[{Taori et~al.(2023)Taori, Gulrajani, Zhang, Dubois, Li, Guestrin,
  Liang, and Hashimoto}]{alpaca}
Rohan Taori, Ishaan Gulrajani, Tianyi Zhang, Yann Dubois, Xuechen Li, Carlos
  Guestrin, Percy Liang, and Tatsunori~B. Hashimoto. 2023.
\newblock Stanford alpaca: An instruction-following llama model.
\newblock \url{https://github.com/tatsu-lab/stanford_alpaca}.

\bibitem[{Team(2023)}]{mpt}
MosaicML~NLP Team. 2023.
\newblock Introducing mpt-7b: A new standard for open-source, commercially
  usable llms.
\newblock Accessed: 2023-05-05.

\bibitem[{Touvron et~al.(2023)Touvron, Lavril, Izacard, Martinet, Lachaux,
  Lacroix, Rozi{\`e}re, Goyal, Hambro, Azhar, Rodriguez, Joulin, Grave, and
  Lample}]{llama}
Hugo Touvron, Thibaut Lavril, Gautier Izacard, Xavier Martinet, Marie-Anne
  Lachaux, Timoth{\'e}e Lacroix, Baptiste Rozi{\`e}re, Naman Goyal, Eric
  Hambro, Faisal Azhar, Aurelien Rodriguez, Armand Joulin, Edouard Grave, and
  Guillaume Lample. 2023.
\newblock \href {https://api.semanticscholar.org/CorpusID:257219404} {Llama:
  Open and efficient foundation language models}.
\newblock \emph{ArXiv}, abs/2302.13971.

\bibitem[{Wan et~al.(2024{\natexlab{a}})Wan, Huang, Cai, Quan, Bi, and
  Shi}]{wan2024knowledge}
Fanqi Wan, Xinting Huang, Deng Cai, Xiaojun Quan, Wei Bi, and Shuming Shi.
  2024{\natexlab{a}}.
\newblock \href {https://openreview.net/pdf?id=jiDsk12qcz} {Knowledge fusion of
  large language models}.
\newblock In \emph{The Twelfth International Conference on Learning
  Representations}.

\bibitem[{Wan et~al.(2024{\natexlab{b}})Wan, Yang, Zhong, Quan, Huang, and
  Bi}]{wan2024fusechat}
Fanqi Wan, Ziyi Yang, Longguang Zhong, Xiaojun Quan, Xinting Huang, and Wei Bi.
  2024{\natexlab{b}}.
\newblock Fusechat: Knowledge fusion of chat models.
\newblock \emph{arXiv preprint arXiv:2402.16107}.

\bibitem[{Wortsman et~al.(2022)Wortsman, Ilharco, Gadre, Roelofs,
  Gontijo-Lopes, Morcos, Namkoong, Farhadi, Carmon, Kornblith, and
  Schmidt}]{soups}
Mitchell Wortsman, Gabriel Ilharco, Samir~Ya Gadre, Rebecca Roelofs, Raphael
  Gontijo-Lopes, Ari~S Morcos, Hongseok Namkoong, Ali Farhadi, Yair Carmon,
  Simon Kornblith, and Ludwig Schmidt. 2022.
\newblock \href {https://proceedings.mlr.press/v162/wortsman22a.html} {Model
  soups: averaging weights of multiple fine-tuned models improves accuracy
  without increasing inference time}.
\newblock In \emph{Proceedings of the 39th International Conference on Machine
  Learning}, volume 162 of \emph{Proceedings of Machine Learning Research},
  pages 23965--23998. PMLR.

\bibitem[{Xu et~al.(2024)Xu, Lu, and Zhang}]{Bridging}
Yangyifan Xu, Jinliang Lu, and Jiajun Zhang. 2024.
\newblock \href {https://openreview.net/forum?id=IOG33p5r6o} {Bridging the gap
  between different vocabularies for {LLM} ensemble}.
\newblock In \emph{2024 Annual Conference of the North American Chapter of the
  Association for Computational Linguistics}.

\bibitem[{Zeng et~al.(2023)Zeng, Liu, Du, Wang, Lai, Ding, Yang, Xu, Zheng, Xia
  et~al.}]{glm}
Aohan Zeng, Xiao Liu, Zhengxiao Du, Zihan Wang, Hanyu Lai, Ming Ding, Zhuoyi
  Yang, Yifan Xu, Wendi Zheng, Xiao Xia, et~al. 2023.
\newblock {GLM-130B:} an open bilingual pre-trained model.
\newblock In \emph{The Eleventh International Conference on Learning
  Representations, {ICLR} 2023, Kigali, Rwanda, May 1-5, 2023}.

\end{thebibliography}

\appendix

% \section{Example Appendix}
% \label{sec:appendix}
% This is an appendix.

\end{document}